\documentclass[journal]{IEEEtran}
\usepackage{amssymb,amsfonts,amsmath,amscd}
\usepackage[ruled,vlined,linesnumbered]{algorithm2e}  %
\usepackage{array}
\usepackage[caption=false,font=footnotesize,labelfont=rm,textfont=rm]{subfig}
\usepackage{textcomp}
\usepackage{stfloats}
\usepackage{url}
\usepackage{verbatim}
\usepackage{graphicx}
\usepackage{subfig}
\usepackage{cite}
\usepackage{hyperref}
\usepackage{multirow}
\usepackage{makecell}
\usepackage{bm}
\hypersetup{
colorlinks=true,
allcolors=blue,
linkcolor=blue,
anchorcolor=blue,
citecolor=blue}
\usepackage{threeparttable}
\usepackage{booktabs}

\usepackage{tabularx}

\DeclareMathAlphabet{\mathscr}{OT1}{pzc}{m}{it}

\begin{document}

\title{ViIK: Flow-based Vision Inverse Kinematics Solver with Fusing Collision Checking}

\author{Qinglong~Meng,
        Chongkun Xia,
        Xueqian Wang*
        \thanks{* Corresponding author}
        \thanks{
        This work was supported by the National Key R\&D Program of China (2022YFB4701400/4701402), 
        National Natural Science Foundation of China (No. U21B6002, 62203260), 
        Guangdong Basic and Applied Basic Research Foundation (2023A1515011773).}
        \thanks{
        Qinglong Meng, and Xueqian Wang are with Tsinghua Shenzhen International Graduate School, Shenzhen 518055, China (e-mail: mengql22@mails.tsinghua.edu.cn; wang.xq@sz.tsinghua.edu.cn;).}
        \thanks{
        Chongkun Xia is with School of Advanced Manufacturing, Sun Yat-sen University, Shenzhen 518107, China (e-mail: xiachk5@mail.sysu.edu.cn;).}}

\markboth{Journal of \LaTeX\ Class Files,~Vol.~14, No.~8, August~2015}%
{Shell \MakeLowercase{\textit{et al.}}: Bare Demo of IEEEtran.cls for IEEE Journals}

\maketitle
\pagestyle{empty}  
\thispagestyle{empty} 

\begin{abstract}
Inverse Kinematics (IK) is to find the robot's configurations that satisfy the target pose of the end effector. In motion planning, diverse configurations were required in case a feasible trajectory was not found. Meanwhile, collision checking (CC), e.g. Oriented bounding box (OBB), Discrete Oriented Polytope (DOP), and Quickhull \cite{quickhull}, needs to be done for each configuration provided by the IK solver to ensure every goal configuration for motion planning is available. This means the classical IK solver and CC algorithm should be executed repeatedly for every configuration. Thus, the preparation time is long when the required number of goal configurations is large, e.g. motion planning in cluster environments. Moreover, structured maps, which might be difficult to obtain, were required by classical collision-checking algorithms. To sidestep such two issues, we propose a flow-based vision method that can output diverse available configurations by fusing inverse kinematics and collision checking, named Vision Inverse Kinematics solver (ViIK). Moreover, ViIK uses RGB images as the perception of environments. ViIK can output 1000 configurations within 40 ms, and the accuracy is about 3 millimeters and 1.5 degrees. The higher accuracy can be obtained by being refined by the classical IK solver within a few iterations. The self-collision rates can be lower than 2\%. The collision-with-env rates can be lower than 10\% in most scenes. The code is available at: \href{https://github.com/AdamQLMeng/ViIK}{https://github.com/AdamQLMeng/ViIK}.
\end{abstract}


\IEEEpeerreviewmaketitle

\begin{figure}[t]
	\centering
	\subfloat{
		\includegraphics[width=0.48\textwidth]{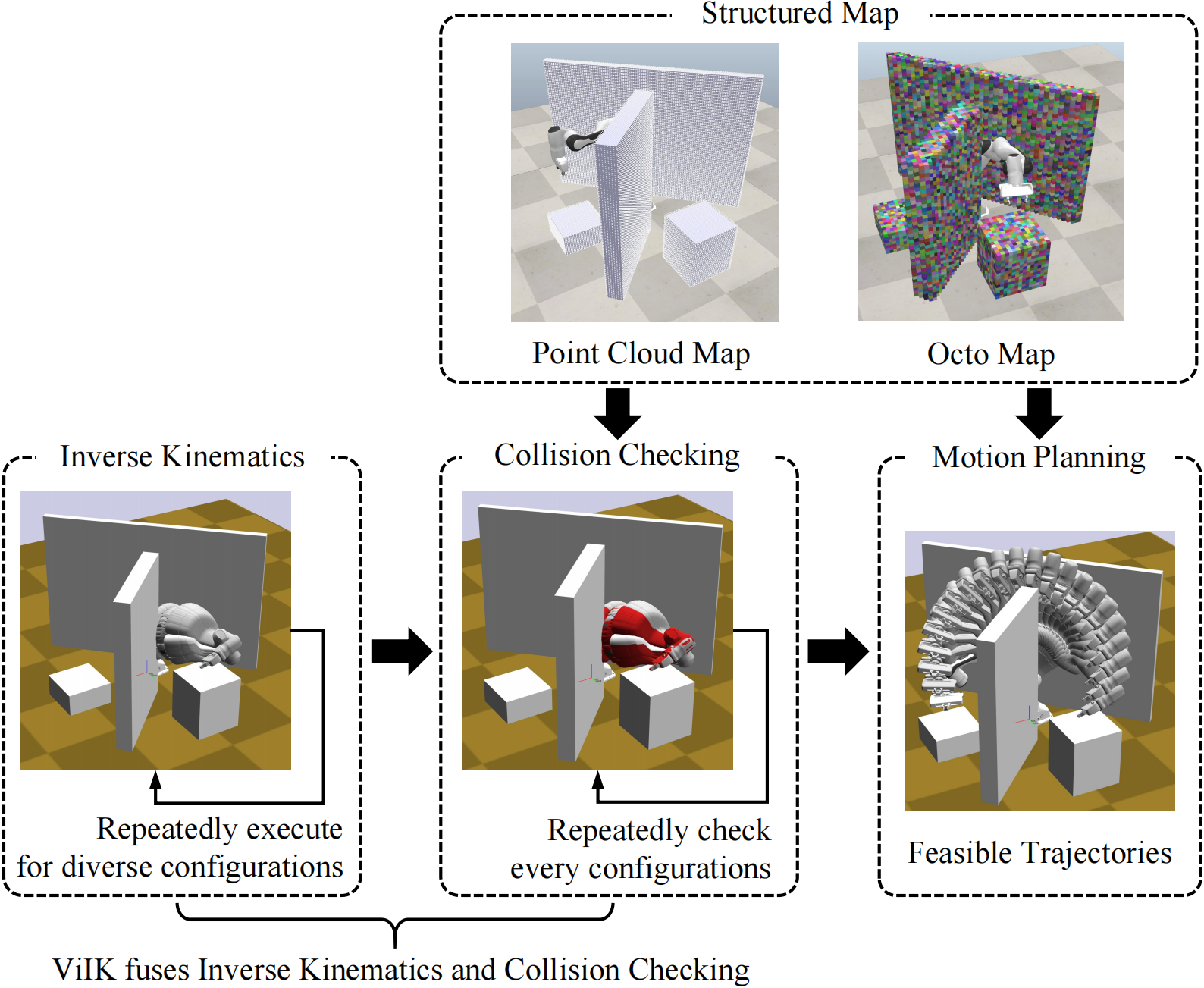}}
    \caption{The architecture of the classical motion planning workflow. The IK solver generates diverse configurations first, then every configuration is checked for collision. (Red configurations are collision-occurred.) At last, all collision-free configurations are used as goal states for motion planning. }
    \label{fig.workflow}
\end{figure}

\section{Introduction}\label{sec.intro}
\IEEEPARstart{I}{verse} Kinematics is a key primitive of high-level decision-making algorithms, e.g. motion planning. For motion planning, the IK solver should output diverse configurations that cover the universal set in case of no feasible goal state or trajectory. Meanwhile, to ensure every goal state is available, all configurations output by the IK solver need collision checking. Then, all collision-free configurations are used as goal states for motion planning. As shown in Fig. \ref{fig.workflow}, in the classical motion planning workflow for a robot with 7+ degrees of freedom (DoF), the classical IK solver and the collision-checking algorithm should be executed repeatedly for every configuration. Therefore, the preparation of motion planning takes a relatively long time when the number of required goal states is large (Fig. \ref{fig.runtime}). Moreover, the classical collision-checking algorithms use structured maps (e.g. point cloud map, and octo map) as the perception of environments.
\par In the prior work, the classical IK solvers include two widely used types: 1) Analytical IK solvers, and 2) Numerical IK solvers. Analytical IK solvers (e.g. IKFast\cite{ikfast}, and IKBT\cite{ikbt}) can output all the solutions for the robots with Dof that is 6 or less. Numerical IK solvers (e.g. TRAC-IK\cite{trac-ik}) can be used for 7+ DoF robots, but only output one solution for a single execution. In the prior learning-based work(\cite{nn-ik-1}-\cite{ikflow-2}), IKFlow\cite{ikflow-1}\cite{ikflow-2} can output diverse configurations in a shorter time compared with the classical IK solver, which benefits from the efficient sampling of Normalizing Flows (NFs). 
\par IK and collision checking as the preparation of motion planning block the subsequent motion planner. Especially for 7+ DoF, classical IK solvers need to be executed repeatedly for every solution as well as collision checking. Thus, a method that can output diverse available configurations directly is needed. To fill this gap, we propose a flow-based vision method that fuses IK and collision checking, named Vision IK solver (ViIK). ViIK can directly output diverse available configurations in various environments, which benefits from the efficient sampling of flow-based models and the use of multi-view RGB images to perceive environments. In this paper, we validate ViIK against the classical method (TRAC-IK+CC) and the learning-based methods (IKFlow and PaddingFlow). The results show our method outputs configurations with competitive position and angular error compared with the learning-based IK solvers. Position and angular error are about 3mm and $1.5^{\circ}$ respectively. The runtime is shorter than the classical method when the number of configurations exceeds 40. The runtime of generating 1000 available configurations is 37ms, while the classical method can be longer than 1s. The self-collision rates are lower than 2\% which is significantly lower than the test set. The collision-with-env rates can be lower than 10\% in most scenes.

\begin{figure*}[t]
	\centering
	\subfloat{
		\includegraphics[width=0.95\textwidth]{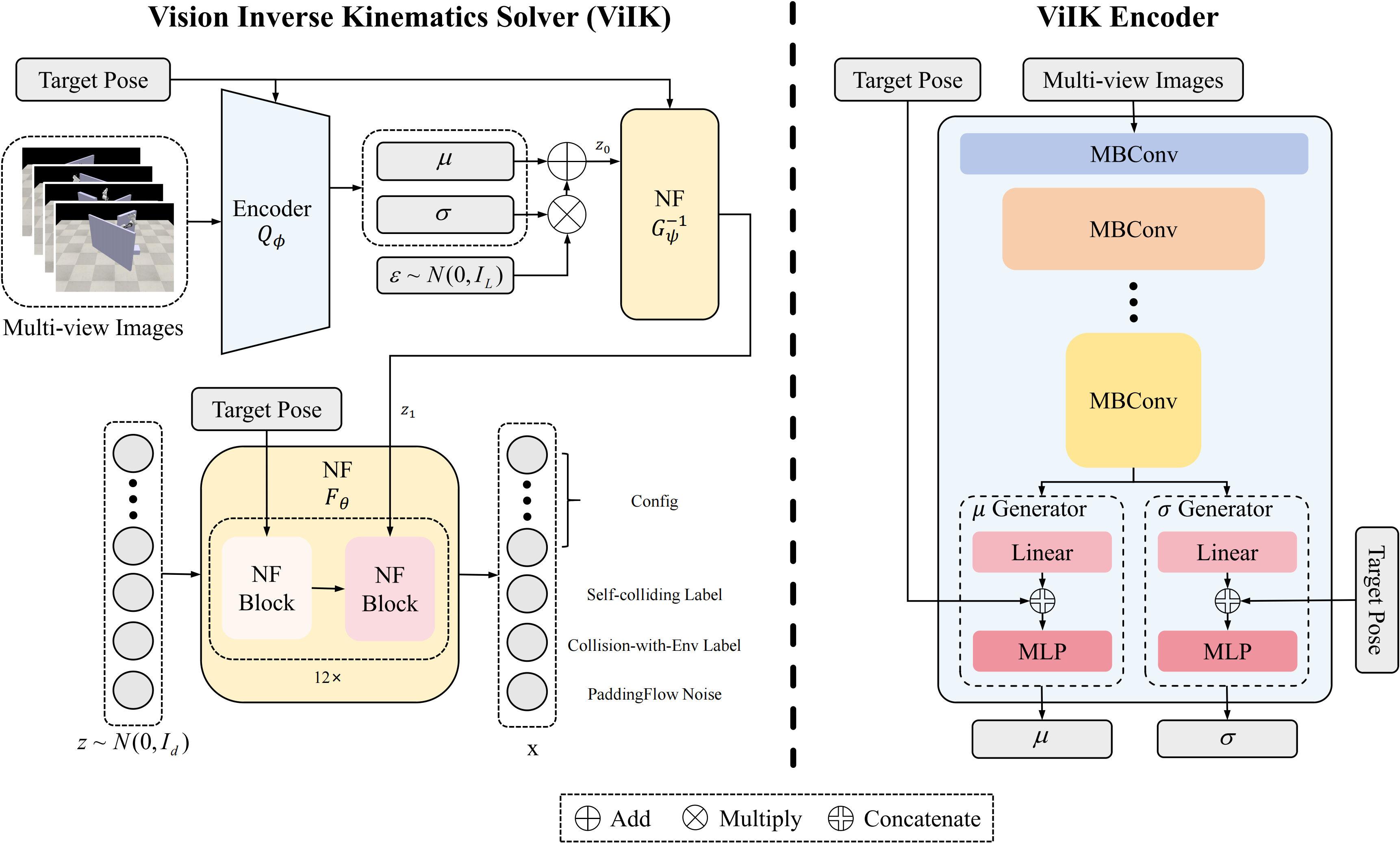}}
    \caption{The architecture of a Vision Inverse Kinematics solver (ViIK-2) in the generative direction. ViIK is a dual flow-based model, one is for generating configurations, and another is for mapping multi-view images into latent space. ViIK Encoder is to map images into latent space with sequential MBConv blocks first, then MLP is used for fusing images and the target pose.}
    \label{fig.model}
\end{figure*}

\section{Background}\label{sec.bg}
In this section, we introduce the prior work of Inverse Kinematics. Moreover, basic concepts of flow-based models are introduced briefly.
\subsection{Inverse Kinematics}\label{sec.ik}
Inverse Kinematics is a primitive of motion planning. To find a feasible trajectory or one satisfying the criterion, motion planners need to use a subset that is large enough to cover the universal set of configurations satisfying the target pose as the goal states. However, the classical IK solvers for 7+ DoF robots (e.g. TRAC-IK) output one configuration by each running. To obtain a large set of configurations, the classical methods need to run enough times with random start states. The learning-based IK solvers \cite{nn-ik-1}-\cite{nn-ik-3} utilize the neural networks for IK, but only can output one configuration for each target pose. \cite{nn-ik-4}-\cite{ikflow-2} can output diverse configurations by using the generative models including Generative Adversarial Networks (GAN), Auto-Encoder (AE), and Normalizing Flows (NFs). IKFlow\cite{ikflow-1}\cite{ikflow-2} has shown the promising results. The accuracy of solutions is within 10mm and  2 degrees. Moreover, the classical IK solver is speeded up significantly by seeding with IKFlow.
\par In practice, the IK solutions need collision checking before being goal states of motion planners, which means the executed times of collision checking might also be large. Therefore, we aim to develop a learning-based method that fuses IK and collision checking to speed up the preparation of motion planning. In this paper, we propose the flow-based Vision Inverse Kinematics solver (ViIK) that can output diverse collision-free solutions to achieve this goal. 
\subsection{Normalizing Flows}\label{sec.nf}
Flow-based models are to model the target distribution as a transformation $F_{\theta}$ of the base distribution that is usually standard normal distribution:
\begin{equation}
x=F_{\theta}(z), \ \mathrm{where} \ z\sim N(0,I).\label{eq.trans}
\end{equation}
Furtherly, the density of $x$ can be obtained by a change of variables:
\begin{equation}
p_{X}(x)=p_{Z}(F_{\theta}^{-1}(x))|J_{F_{\theta}^{-1}}(x)|.\label{eq.density}
\end{equation}
Moreover, the density of $x$ conditioned on $y$ can be written as:
\begin{equation}
p_{X}(x|y)=p_{Z}(F_{\theta}^{-1}(x;y))|J_{F_{\theta}^{-1}}(x|y)|.\label{eq.cond_density}
\end{equation}
\par In practice, we often construct a neural network to fit the transformation $F_{\theta}$. The corresponding objective function is usually Kullback-Leibler (KL) divergence to minimize divergence between the flow-based model $p_{X}(x|y;\theta)$ and the target distribution $p_{X}^{*}(x|y)$, which can be estimated by Monte Carlo\cite{survey_nf}:
\begin{equation}
\begin{aligned}
\mathcal{L}(\mathcal{X},\mathcal{Y};\theta)
\approx-\frac{1}{N}\sum_{i=1}^{N}[&\mathrm{log}p_{Z}(F_{\theta}^{-1}(x^{(i)};y^{(i)}))\\
&+\mathrm{log|det}J_{F_{\theta}^{-1}}(x^{(i)}|y^{(i)})|],
\end{aligned}\label{eq.cond_loss}
\end{equation}
where $\mathcal{X}=\{x^{(i)}\}_{i=1}^{N}$, $\mathcal{Y}=\{y^{(i)}\}_{i=1}^{N}$. 

\begin{figure*}[t]
	\centering
    \captionsetup[subfloat]{labelsep=none,format=plain}
	\subfloat[ TRAC-IK in Env-1\label{fig.examples.e1_trac}]{
		\includegraphics[width=0.23\textwidth]{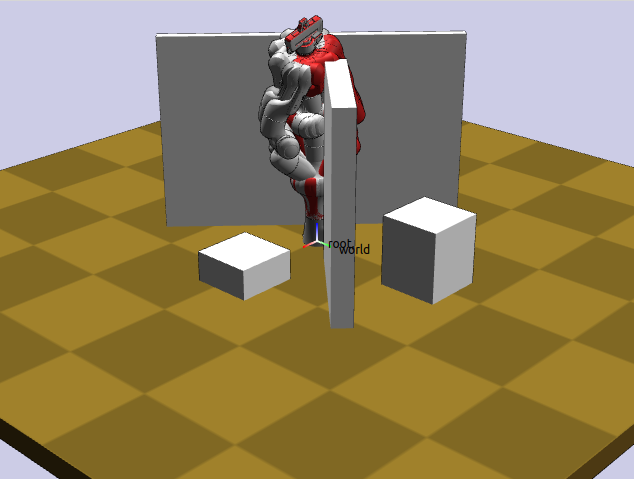}}
    \subfloat[ ViIK in Env-1\label{fig.examples.e1_viik}]{
        \includegraphics[width=0.23\textwidth]{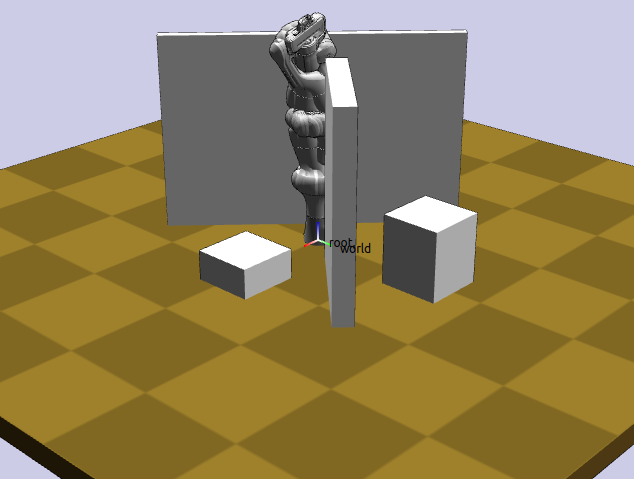}}
    \subfloat[ TRAC-IK in Env-3\label{fig.examples.e3_trac}]{
        \includegraphics[width=0.23\textwidth]{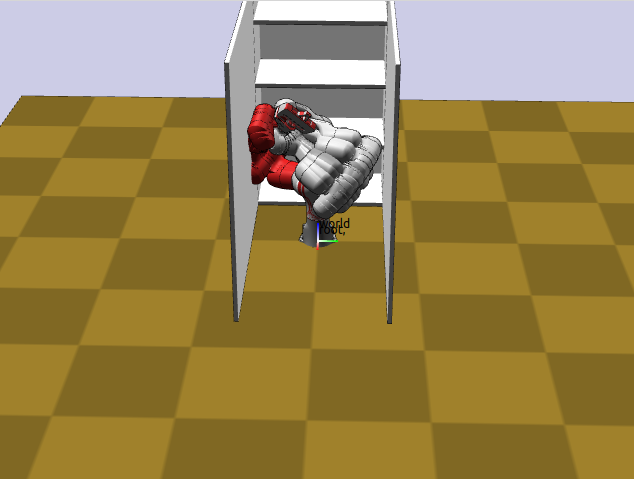}}
    \subfloat[ ViIK in Env-3\label{fig.examples.e3_viik}]{
        \includegraphics[width=0.23\textwidth]{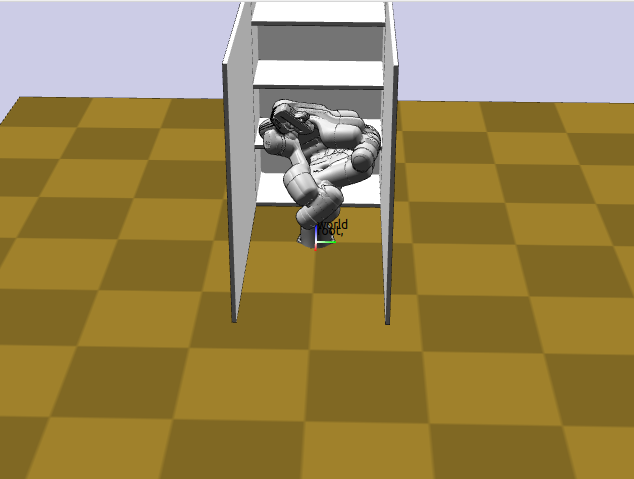}}
    \caption{Examples for comparison of ViIK and the classical IK solver (TRAC-IK). (a), and (b) is an example in Env-1. (c), and (d) is an example in Env-3. Env-1 and Env-3 are the hard scenes among the scenes used in this paper, which have collision rates that are higher than 75\%. The examples show that the solutions output by ViIK have a lower collision rate compared with TRAC-IK. }
    \label{fig.examples}
\end{figure*}

\section{Method}\label{sec.method}
In this section, we introduce the target distribution for approximating the collision-free IK solutions first. Secondly, we introduce the architecture of the base model of ViIK, named VisionFlow. Thirdly, we introduce the dequantization method chosen to sidestep two issues of NF: 1) the discrete data, and 2) the target distribution is a manifold.

\subsection{Target Distribution}\label{sec:dist}
For 7+ DoF robots, the IK solvers output the samples in the sub-space of the configuration space (C-space) where the target pose ($p$) is the condition. We can, firstly, use a conditional distribution to approximate the sub-space, where the density can be denoted as $p_{C}(c|p)$. C-space can also be divided into two sub-spaces: collision-free space denoted as $C_{c-f}$ and collision space denoted as $C_{c}$. Further, conditional distribution can be used for collision checking, where the density can be denoted as $p_{C}(c|x_{c})$, where $x_{c}$ is whether collision-with-env occurs. Therefore, the collision-free configurations satisfying the target pose can be approximated by the two conditional distributions, denoted as:
\begin{equation}
\left\{
\begin{array}{ll}
p_{1}=p_{C_{c-f}}(c|p)   &\\[2pt]
p_{2}=p_{C}(c|p,x_{c})   &\\[2pt]
p_{3}=p_{X'}(x'|p)
\end{array}
\right.
,
\end{equation}
where  $x'=(c,x_{c})$, $x_{c}$ is whether collision-with-env occurs. The first one $p_{C_{c-f}}(c|p)$ is to train models without negative samples where models might perform badly near the boundary. 
As for the second one $p_{C}(c|p,x_{c})$, the model has similar performance on position and angular error compared with $p_{X}((c,x_{c})|p)$, but performs badly on collision-sith-env and self-collision rates, as shown in Tab. \ref{tab.dist}. In this paper, we use $p_{X}((c,x_{c})|p)$ as the target distribution. After adding self-collision in the same way, the target distribution is:
\begin{equation}
p_{X}(x|p),
\end{equation}
where $x=(c,x_{s-c},x_{c})$, $x_{s-c}$ is whether self-collision occurs. Moreover, to output collision-free configurations in different environments and perceive environments with RGB images, the multi-view images ($\mathcal{I}$) are condition, where the density is written as:
\begin{equation}
p_{X}(x|p, \mathcal{I}),\label{eq.dist}
\end{equation}
where $\mathcal{I}=\{I_{j}\}_{j=1}^{M}$, $M$ denotes the number of view.

\subsection{VisionFlow}\label{sec:visionflow}
Due to the large number of configurations required for motion planning, the efficient sampling of NF is suitable for IK. Moreover, for perceiving environments without structured maps, the RGB images need to be the input of NF. Therefore, we propose a Vision-conditioned flow-based model, named VisionFlow, as shown in Fig. \ref{fig.model}. VisionFlow is a flow-based model that takes instruction (It's the target pose in ViIK) and RGB images as conditions as shown in Eq. \ref{eq.dist}.
\subsubsection{Flow-based Encoder of VisionFlow}
To fuse instruction and images, the flow-based encoder of VisionFlow has a conv-based encoder and a conditional normalizing flow. The conv-based encoder can be written as:
\begin{equation}
\left\{
\begin{array}{ll}
\mu \ (\mathrm{or} \ \sigma) = \mathrm{MLP}([p, \mathrm{Linear}(\mathrm{MBConv}(\mathcal{I}))])   &\\[2pt]
z_{0}=\mathrm{Reparameterize}(\mu, \sigma)   
\end{array}
\right.
,
\end{equation}
where $z_{0}$ obeys the Gaussian distribution $N(\mu, \sigma^{2}I)$, and is the latent representation of the environment conditioned on the target pose $p$. Further, the conditional normalizing flow normalizes the latent distribution:
\begin{equation}
z_{1}=G_{\psi}^{-1}(z_{0};p).
\end{equation}
Thus, the density of the latent distribution can be written as:
\begin{equation}
\begin{aligned}
p_{Z_{1}}(z_{1}|p, \mathcal{I})
&=p_{Z_{0}}(z_{0}|p, \mathcal{I})|J_{G_{\psi}}(z_{0}|p)|\\[2pt]
&= Q_{\phi}(z_{0}|p, \mathcal{I})|J_{G_{\psi}}(z_{0}|p)|.
\end{aligned}\label{eq.encoder}
\end{equation}
\subsubsection{Loss Function}
In the normalizing direction, the density of the target distribution (Eq. \ref{eq.dist}) can be written as:
\begin{equation}
p_{X}(x|p, z_{1})=p_{Z}(F_{\theta}^{-1}(x;p, z_{1}))|J_{F_{\theta}^{-1}}(x|p, z_{1})|.\label{eq.gen_dir_tr}
\end{equation}
Because the prior distributions of the target pose $p$ and the multi-view images $\mathcal{I}$ are fixed, regulating the latent distribution $p_{Z_{1}}(z_{1}|p, \mathcal{I})$ is not necessary. Thus, we use the maximum log-likelihood of $x$ (Eq. \ref{eq.cond_loss}) as the loss function of VisionFlow:
\begin{equation}
\begin{aligned}
\mathcal{L}(\mathcal{X}, \mathcal{P}, \mathcal{Z}_{1};\theta, \psi, \phi)&=-\frac{1}{B}\sum_{i=1}^{B}[\mathrm{log}p_{Z}(F_{\theta}^{-1}(x^{(i)};p^{(i)}, z_{1}^{(i)}))\\
&+\mathrm{log|det}J_{F_{\theta}^{-1}}(x^{(i)}|p^{(i)}, z_{1}^{(i)})|].
\end{aligned}
\end{equation}
To reduce the cost of GPU memory and training time, only one set of multi-view images is used for each batch. It's equivalent to solving $B$ IK problems of the same environment in each training step that the batch size is $B$. The loss function can be written as:
\begin{equation}
\begin{aligned}
\mathcal{L}(\mathcal{X}, \mathcal{P}, z_{1};\theta, \psi, \phi)&=-\frac{1}{B}\sum_{i=1}^{B}[\mathrm{log}p_{Z}(F_{\theta}^{-1}(x^{(i)};p^{(i)}, z_{1})\\
&+\mathrm{log|det}J_{F_{\theta}^{-1}}(x^{(i)}|p^{(i)}, z_{1})|].\label{eq.viik_loss}
\end{aligned}
\end{equation}

\begin{table*}[t]
    \caption{Target distributions comparison on the test set (150,000 random IK problems of Panda manipulator).}
    \centering
    \begin{threeparttable}
    \begin{tabular}{clcccc}
        \toprule [1pt]\noalign{\vskip 2pt}
        & Model & Position Error (mm, $\downarrow$) & Angular Error (deg, $\downarrow$) & Collision-with-Env Rates (\%, $\uparrow$) & Self-collision Rates (\%, $\uparrow$)\\[5pt]
        \hline\noalign{\vskip 5pt}
        \multirow{4}{*}{Env-1}
        &$p_{2}$     & 3.85 & 4.64 & 74.2 & 4.00 \\
        &$p_{3}$     & 3.48 & 3.24 & 18.7 & 1.88 \\[2pt]
        \cmidrule(l){2-6} \noalign{\vskip 1pt}
        &Test Set      & - & - & 75.1 & 5.42 \\[4pt]
        \hline\noalign{\vskip 5pt}
        \multirow{4}{*}{Env-2}
        &$p_{2}$     & 3.85 & 4.64 & 32.5 & 4.00 \\
        &$p_{3}$     & 3.39 & 3.20 & 1.58 & 1.87 \\[2pt]
        \cmidrule(l){2-6} \noalign{\vskip 1pt}
        &Test Set      & - & - & 32.8 & 5.42 \\[4pt]
        \bottomrule [1pt]
    \end{tabular}
    \end{threeparttable}
\label{tab.dist}
\end{table*}

\subsection{Dequantization}\label{sec:dequan}
Issues that degrade the performance of flow-based models include: 1) the discrete data \cite{discrete}; and 2) the target distribution is a manifold \cite{softflow}. For the first issue, the discrete data, datasets are sets of samples from the true distribution. The sampling process can be seen as quantization. Training a flow-based model on the discrete data is to fit a continuous density model to a discrete distribution, which might lead the model to collapse into a degenerate mixture of point masses\cite{discrete}. As shown in Eq. \ref{eq.dist}, the target distribution might be a manifold in the C-space by approximating IK solutions by a conditional distribution.
\par Two abovementioned issues can be sidestepped by dequantization. In this paper, we choose to implement the dequantization method, PaddingFlow\cite{paddingflow}, which is widely suitable and easy to implement. By referencing the results of IK experiments in PaddingFlow\cite{paddingflow} we modify the PaddingFlow noise for the data used in this paper. Unlike the configuration, the self-collision and collision-with-env flags are boolean values. Thus, the Gaussian noise is added to the self-collision flag ($x_{sc}$) and collision-with-env flag ($x_{c}$):
\begin{equation}
\left\{
\begin{array}{ll}
\varepsilon_{2}\sim N(0,a^{2}I_{2}) &\\[2pt]
\varepsilon\sim N(0,b^{2}) &\\[2pt]
x'= (c, (x_{s-c}, x_{c})+\varepsilon_{2},\varepsilon)&\\
\end{array}
\right.
,
\end{equation}
where $a$, and $b$ denote the variances of data noise, and padding-dimensional noise respectively.

\section{Experiments}\label{sec.exp}
In this section, we evaluate ViIK against the classical and learning-based methods. Firstly, we test runtime compared to TRAC-IK+CC. Secondly, we evaluate the position and angular errors against the learning-based methods (IKFlow and PaddingFlow). Thirdly, we present the self-collision and collision-with-env rates in 10 typical scenes \cite{benchmark}.
\par The scenes used in this paper contain various environments, including Env-2, and Env-6 -- Env-10 are low cluttered, Env-1, Env-3, and Env-4 are medium cluttered, and Env-5 is high cluttered. Moreover, Env-4 is a complex environment with many narrow passages, and Env-2, and Env-6 -- Env-9 are trivial environments with narrow passages. As for Env-10, it is a trivial environment with no narrow passage.

\subsection{Target Distributions Comparison}\label{sec.exp.dist}
We compare two target distributions ($p_{2}$, and $p_{3}$) on position error, angular error, collision-with-env rates, and self-collision rates. The results (Tab. \ref{tab.dist}) show $p_{2}$ performs better on all 4 metrics. Especially on collision-with-env and self-collision rates, $p_{3}$ performs badly, even similarly to the test set.

\begin{figure}[ht]
	\centering
    \captionsetup[subfloat]{labelsep=none,format=plain}
	\subfloat[\label{fig.runtime.viik_classical}]{
		\includegraphics[width=0.22\textwidth]{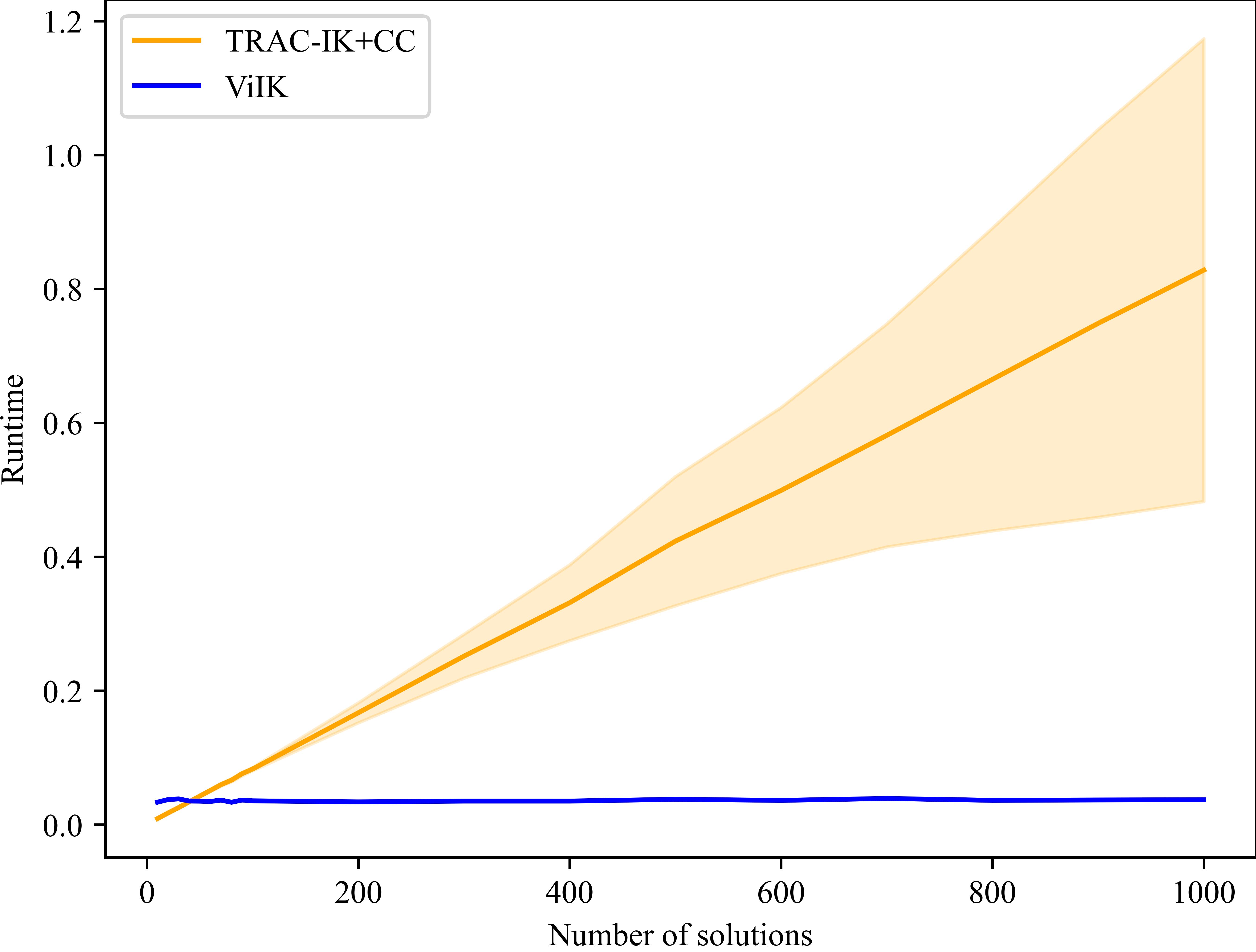}}
    \subfloat[\label{fig.runtime.classical}]{
        \includegraphics[width=0.22\textwidth]{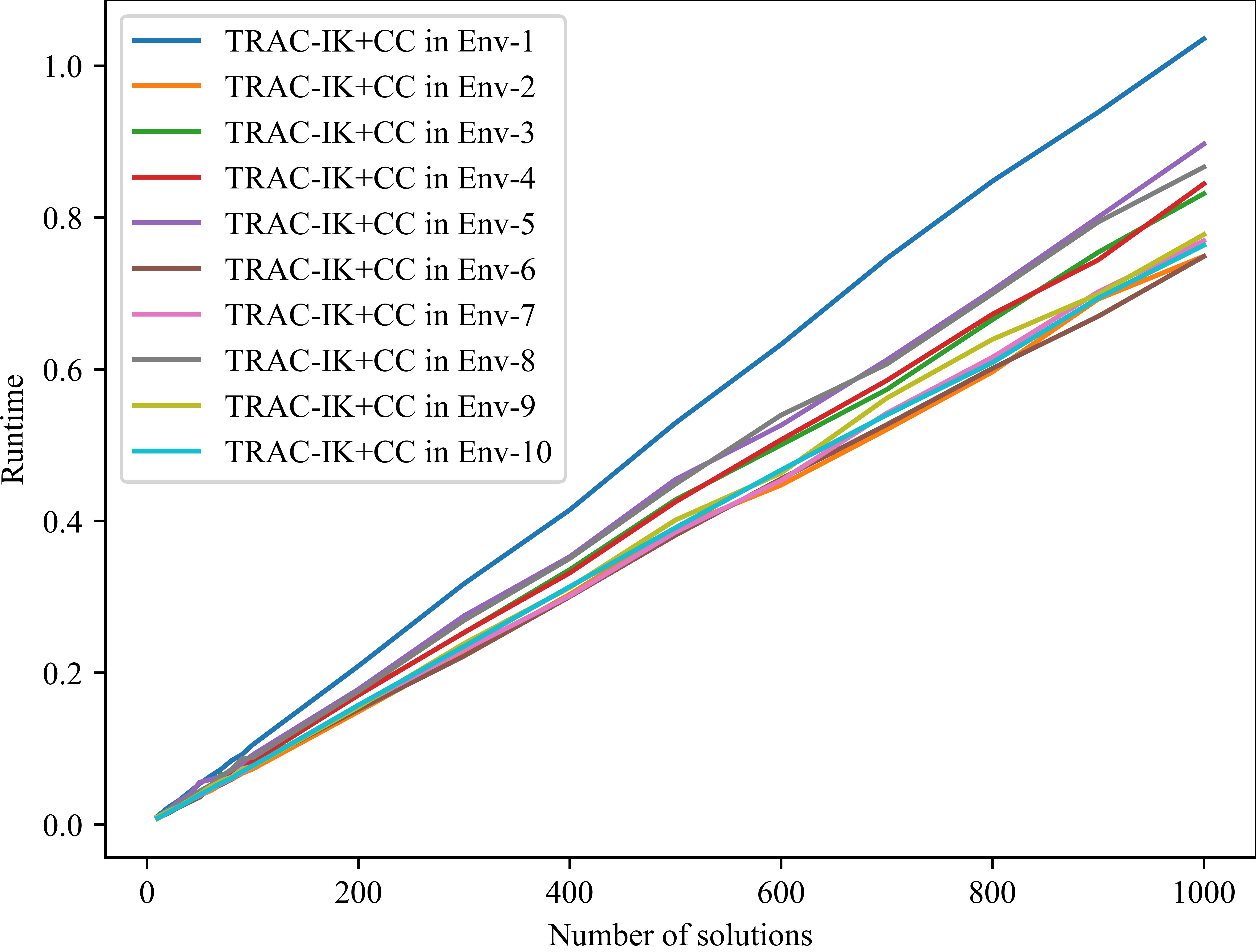}}\\[-2pt]
    \subfloat[\label{fig.runtime.local}]{
		\includegraphics[width=0.22\textwidth]{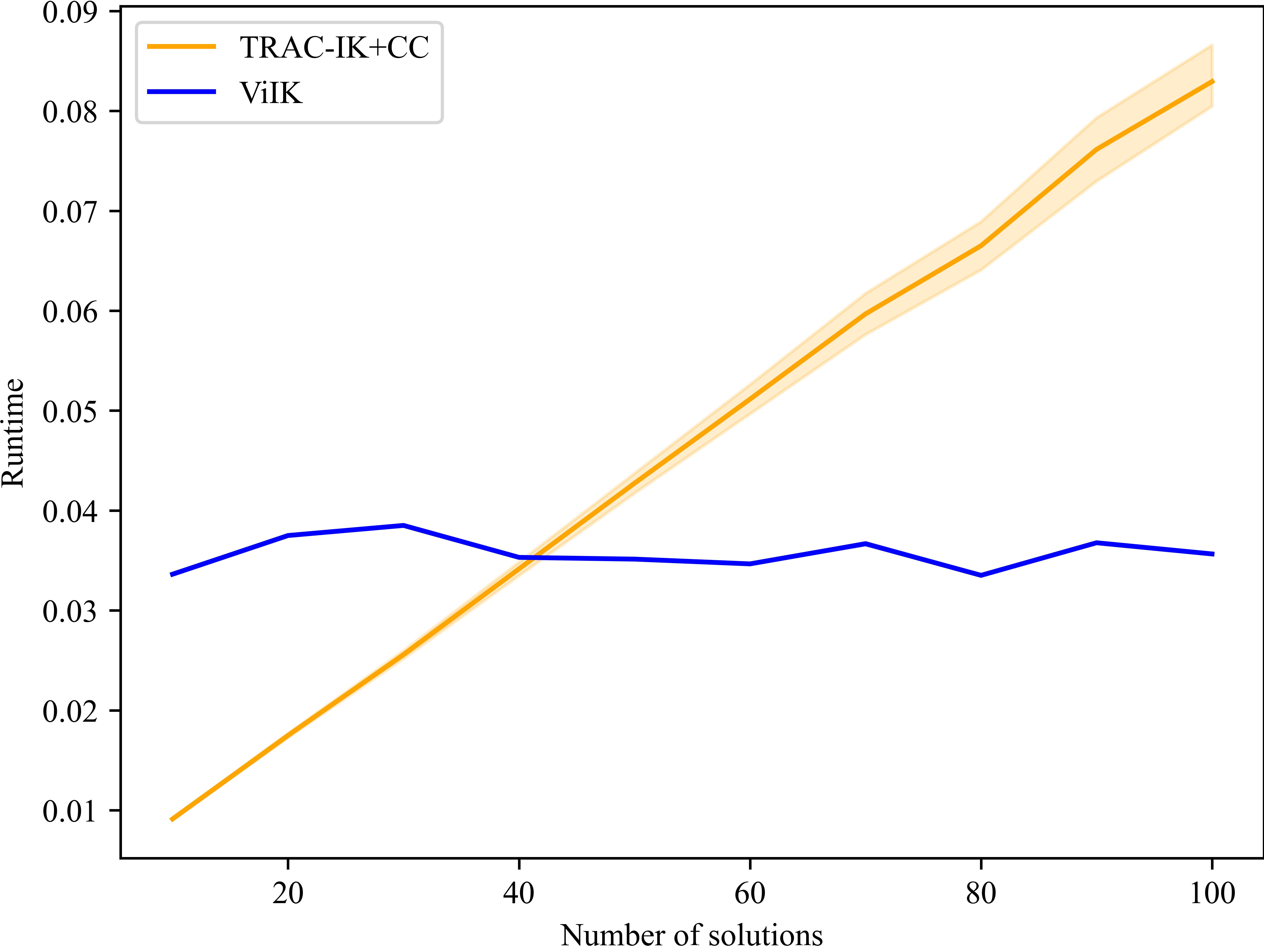}}
    \subfloat[\label{fig.runtime.viik}]{
        \includegraphics[width=0.22\textwidth]{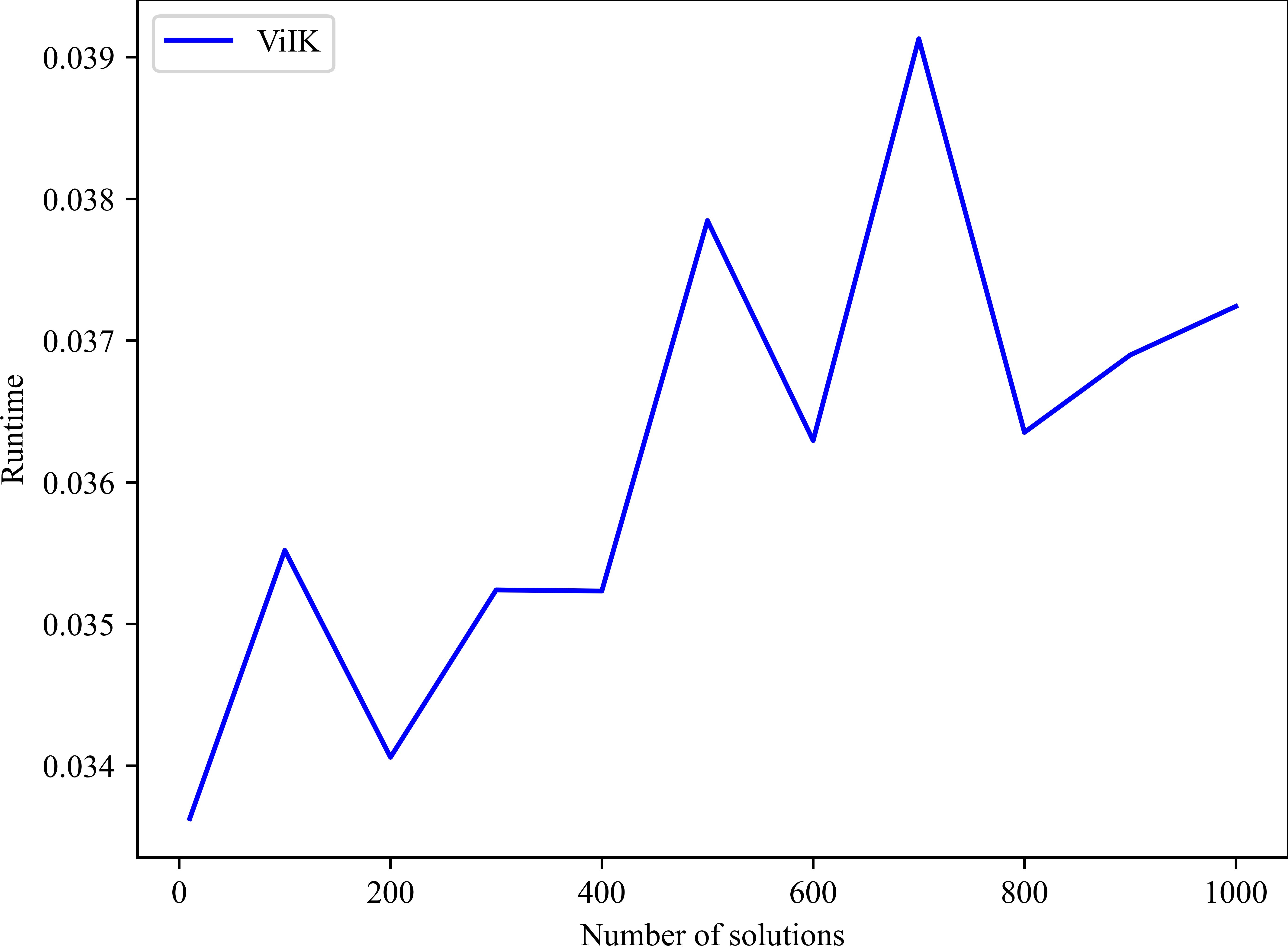}}
    \caption{The runtime of ViIK-2, and TRAC-IK+CC. The tolerance of TRAC-IK is set to $1\times 10^{-3}$, similar to the errors of ViIK. Collision checking in TRAC-IK+CC uses mesh colliders.}
    \label{fig.runtime}
\end{figure}

\subsection{Runtime Compared with the Classical Method}\label{sec.exp.runtime}
We compare the runtime of ViIK to the classical method, TRAC-IK+CC. The tolerance of TRAC-IK is set to $1\times 10^{-3}$ which is similar to the errors of ViIK shown in Tab. \ref{tab.error}. Moreover, Mesh Collider is used for collision checking in TRAC+CC. We let TRAC-IK repeatedly execute until the number of output solutions is the same as ViIK, $t_{ik}$ denotes the total runtime of TRAC-IK. Then we run collision checking for every configuration output by TRAC-IK, $t_{cc}$ denotes the total runtime of collision checking. The runtime of TRAC-IK+CC is:
\begin{equation}
t_{classical}=t_{ik}+t_{cc}
\end{equation}
\par In Fig. \ref{fig.runtime}, we present the runtime of different numbers of solutions from 10 to 1000. The results show that ViIK outperforms TRAC-IK when the number of solutions exceeds 40. The runtime of 100 and 1000 solutions is about 1/2 and 1/17 compared to the classical method.

\begin{table*}[t]
    \caption{Position and Angular Errors on the test set (150,000 random IK problems of Panda manipulator).}
    \centering
    \begin{threeparttable}
    \resizebox{\linewidth}{!}{
    \begin{tabular}{clccccccccccc}
        \toprule [1pt]\noalign{\vskip 2pt}
        & Model & Env-1 & Env-2 & Env-3 & Env-4 & Env-5 & Env-6 & Env-7 & Env-8 & Env-9 & Env-10 & Mean\\[5pt]
        \hline\noalign{\vskip 5pt}
        \multirow{6}{*}{Position Error (mm, $\downarrow$)}
        &IKFlow        & - & - & - & - & - & - & - & - & - & - & 6.86 \\
        &PaddingFlow   & - & - & - & - & - & - & - & - & - & - & 5.96 \\[2pt]
        \cmidrule(l){2-13} \noalign{\vskip 1pt}
        &ViIK-2      & 3.48 & 3.39 & 4.25 & 4.26 & 4.22 & 4.22 & 4.63 & 4.62 & 4.50 & 4.50 & 4.21 \\
        &ViIK-5      & 3.08 & 3.23 & 3.06 & 3.06 & 3.06 & 4.01 & 4.01 & 4.00 & 4.01 & 4.01 & \textbf{3.55} \\
        &ViIK-10     & 5.16 & 4.93 & 5.03 & 5.02 & 5.20 & 4.97 & 4.94 & 4.97 & 4.98 & 4.95 & 5.02 \\[2pt]
        \hline\noalign{\vskip 5pt}
        \multirow{6}{*}{Angular Error (deg, $\downarrow$)}
        &IKFlow        & - & - & - & - & - & - & - & - & - & - & 2.38 \\
        &PaddingFlow   & - & - & - & - & - & - & - & - & - & - & \textbf{0.62} \\[2pt]
        \cmidrule(l){2-13} \noalign{\vskip 1pt}
        &ViIK-2      & 3.24 & 3.20 & 4.37 & 4.37 & 4.44 & 4.45 & 3.90 & 3.90 & 4.12 & 4.12 & 4.01 \\
        &ViIK-5      & 1.47 & 1.52 & 1.47 & 1.47 & 1.46 & 1.53 & 1.53 & 1.53 & 1.53 & 1.53 & 1.51 \\
        &ViIK-10     & 1.58 & 1.54 & 1.56 & 1.56 & 1.58 & 1.55 & 1.54 & 1.55 & 1.55 & 1.55 & 1.56 \\[2pt]   
        \bottomrule [1pt]
    \end{tabular}
    }
    \end{threeparttable}
\label{tab.error}
\end{table*}

\subsection{Position Error and Angular Error Compared with the Learning-based Methods}\label{sec.exp.error}
We evaluate ViIK on the test set with 150,000 random IK problems of the Panda manipulator against IKFlow and PaddingFlow. We let ViIK, IKFlow, and PaddingFlow output 1000 solutions for each target pose. In Tab. \ref{tab.error}, three models (ViIK-2, ViIK-5, and ViIK-10 trained on the data of 2, 5, and 10 scenes respectively) can output competitive solutions compared with existing learning-based methods. ViIK outperforms IKFlow and PaddingFlow on position error which is about 1/2 of IKFlow, but on angular error, ViIK performs worse than PaddingFlow.

\begin{table*}[t]
    \caption{Self-collision and Collision-with-Env Rates on the test set (10,000 random IK problems of Panda manipulator).}
    \centering
    \begin{threeparttable}
    \resizebox{\linewidth}{!}{
    \begin{tabular}{clccccccccccc}
        \toprule [1pt]\noalign{\vskip 2pt}
        & Model & Env-1 & Env-2 & Env-3 & Env-4 & Env-5 & Env-6 & Env-7 & Env-8 & Env-9 & Env-10 & Mean\\[5pt]
        \hline\noalign{\vskip 5pt}
        \multirow{5}{*}{Collision-with-Env (\%, $\downarrow$)}
        &ViIK-2      & 18.7 & \textbf{1.58} & \textbf{10.8} & \textbf{10.8} & 29.7 & \textbf{4.00} & 18.6 & 16.0 & 21.6 & 21.7 & 15.3 \\
        &ViIK-5      & \textbf{17.1} & 2.16 & 11.2 & 11.2 & \textbf{27.5} & 13.1 & 21.8 & 19.9 & 28.9 & 26.3 & 17.9\\
        &ViIK-10     & 27.0 & 2.34 & 17.0 & 17.0 & 41.7 & 4.37 & \textbf{5.60} & \textbf{6.18} & \textbf{9.04} & \textbf{7.50} & \textbf{13.8} \\[2pt]
        \cmidrule(l){2-13} \noalign{\vskip 1pt}
        &Test Set      & 75.1 & 32.8 & 76.6 & 76.7 & 87.3 & 26.5 & 35.7 & 33.4 & 38.1 & 40.2 & 52.2 \\[4pt]
        \hline\noalign{\vskip 5pt}
        \multirow{3}{*}{IoU of Collision-with-Env (\%, $\uparrow$)}
        &ViIK-2      & 69.2 & \textbf{96.7} & \textbf{80.3} & \textbf{80.3} & 52.4 & \textbf{92.2} & 69.7 & 71.0 & 64.1 & 64.8 & 74.1 \\
        &ViIK-5      & \textbf{71.6} & 95.7 & 79.9 & 79.8 & \textbf{59.4} & 69.1 & 64.8 & 65.5 & 58.2 & 62.5 & 70.7 \\
        &ViIK-10     & 59.2 & 94.4 & 71.4 & 71.3 & 44.1 & 90.8 & \textbf{87.9} & \textbf{87.8} & \textbf{82.2} & \textbf{85.0} & \textbf{77.4} \\[4pt]   
        \hline\noalign{\vskip 5pt}
        \multirow{5}{*}{Self-Collision (\%, $\downarrow$)}
        &ViIK-2      & 1.88 & 1.87 & 1.94 & 1.93 & 2.05 & 2.05 & 1.85 & 1.86 & 1.83 & 1.82 & 1.91 \\
        &ViIK-5      & 1.52 & 1.52 & 1.51 & 1.52 & 1.52 & 1.87 & 1.86 & 1.86 & 1.87 & 1.87 & \textbf{1.69} \\
        &ViIK-10     & 1.85 & 1.90 & 1.86 & 1.86 & 1.84 & 1.87 & 1.90 & 1.88 & 1.88 & 1.88 & 1.87 \\[2pt]
        \cmidrule(l){2-13} \noalign{\vskip 1pt}
        &Test Set      & - & - & - & - & - & - & - & - & - & - & 5.42 \\[4pt]
        \hline\noalign{\vskip 5pt}
        \multirow{3}{*}{IoU of Self-collision (\%, $\uparrow$)}
        &ViIK-2      & 96.7 & 96.7 & 96.4 & 96.4 & 96.1 & 96.1 & 96.7 & 96.7 & 96.7 & 96.7 & 96.5 \\
        &ViIK-5      & 97.3 & 97.2 & 97.3 & 97.3 & 97.3 & 96.5 & 96.5 & 96.5 & 96.5 & 96.5 & \textbf{96.9} \\
        &ViIK-10     & 96.3 & 96.4 & 96.3 & 96.3 & 96.3 & 96.4 & 96.4 & 96.4 & 96.4 & 96.4 & 96.4 \\[2pt]   
        \bottomrule [1pt]
    \end{tabular}
    }
    \end{threeparttable}
\label{tab.collision}
\end{table*}

\subsection{Self-collision and Collision-with-Env Rates}\label{sec.exp.collision}
We present ViIK rates of self-collision and collision with the environments on the test sets. We test self-collision and collision-with-env rates on the test set with 10,000 random IK problems of the Panda manipulator. For each target pose, ViIK outputs 1000 solutions. For different scenes, we evaluate ViIK on the same test set but with different images of each scene. We report self-collision and collision-with-env rates over 10,000 target poses, as shown in Tab. \ref{tab.collision}. The self-collision rates of ViIK are lower than 2\%, which is significantly lower than the test set. As for the collision-with-env rates, ViIK can perform well in most scenes but is not ideal in some hard scenes (Env-1, and Env-5).

\section{Conclusion}\label{sec.conclusion}
In this paper, to achieve an IK solver that can output diverse collision-free configurations, we propose a flow-based vision method, which fuses Inverse Kinematics and collision checking, named Vision Inverse Kinematics solver (ViIK). ViIK outputs diverse available configurations using RGB multi-view images as the perception of environments. The results show that ViIK can output 1000 configurations in 40ms which is about 1/17 compared to TRAC-IK+CC. ViIK can output competitive solutions for position and angular errors compared to existing learning-based methods. Moreover, self-collision rates are significantly lower than the test set and can be lower than 2\%. Collision-with-env rates are significantly lower than the test set, which are lower than 10\% in most scenes.
\begin{figure*}[t]
	\centering
    \captionsetup[subfloat]{labelsep=none,format=plain,labelformat=empty}
	\subfloat[Env-1\label{fig.benchmark.e1}]{
		\includegraphics[width=0.18\textwidth]{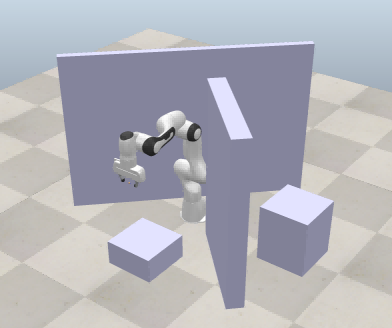}}
    \subfloat[Env-2\label{fig.benchmark.e2}]{
        \includegraphics[width=0.18\textwidth]{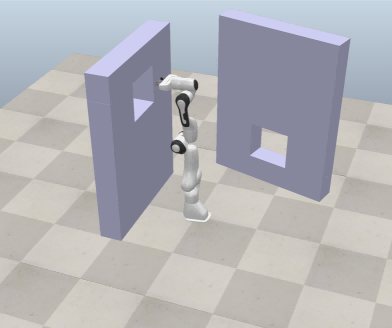}}
    \subfloat[Env-3\label{fig.benchmark.e3}]{
        \includegraphics[width=0.18\textwidth]{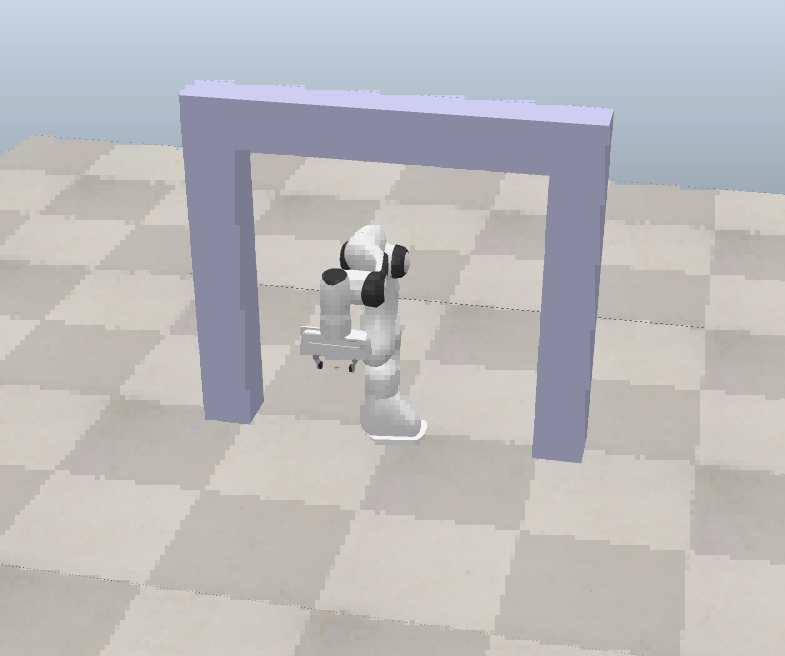}}
    \subfloat[Env-4\label{fig.benchmark.e4}]{
        \includegraphics[width=0.18\textwidth]{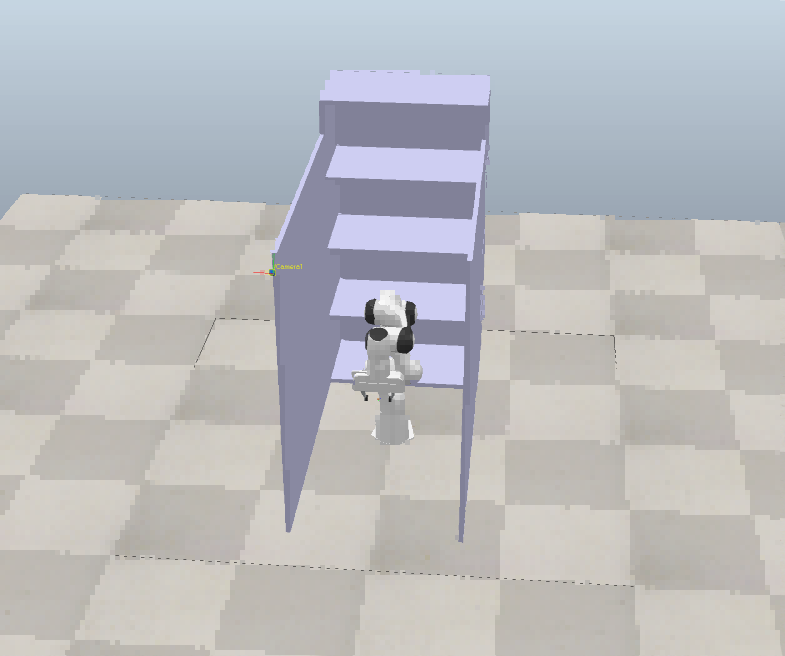}}
    \subfloat[Env-5\label{fig.benchmark.e5}]{
        \includegraphics[width=0.18\textwidth]{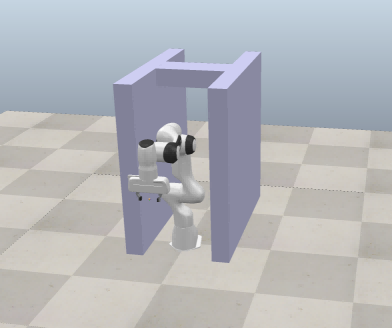}}\\[-4pt]
    \subfloat[Env-6\label{fig.benchmark.e6}]{
		\includegraphics[width=0.18\textwidth]{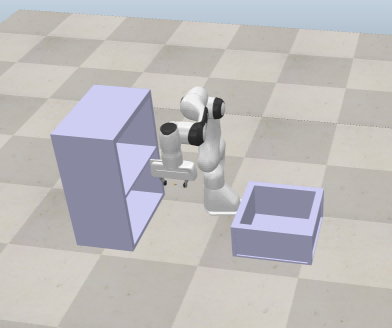}}
    \subfloat[Env-7\label{fig.benchmark.e7}]{
        \includegraphics[width=0.18\textwidth]{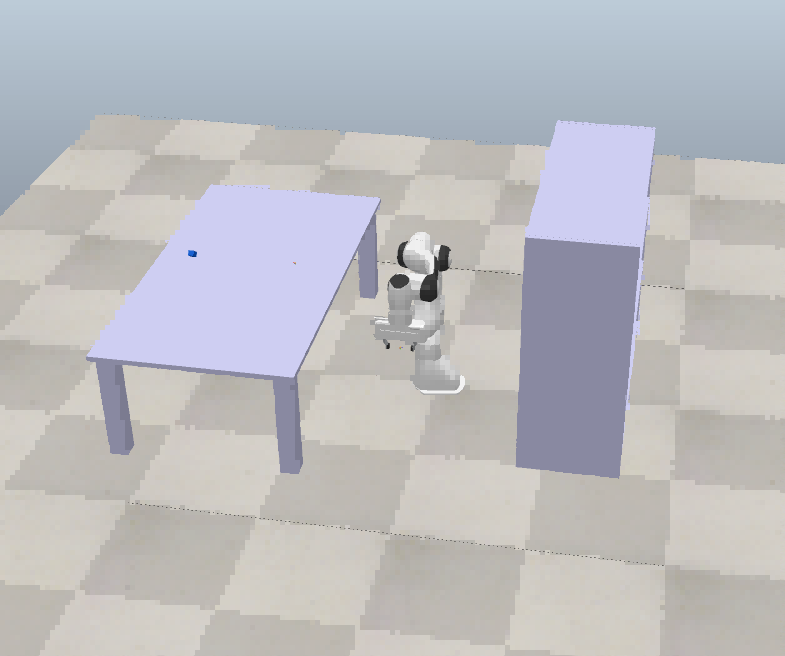}}
    \subfloat[Env-8\label{fig.benchmark.e8}]{
        \includegraphics[width=0.18\textwidth]{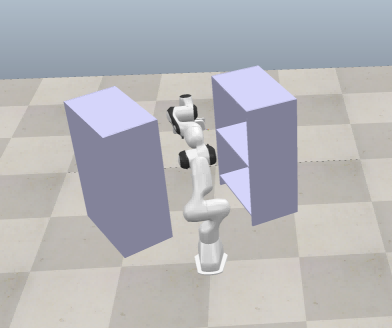}}
    \subfloat[Env-9\label{fig.benchmark.e9}]{
        \includegraphics[width=0.18\textwidth]{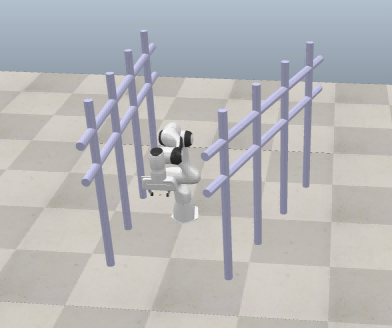}}
    \subfloat[Env-10\label{fig.benchmark.e10}]{
        \includegraphics[width=0.18\textwidth]{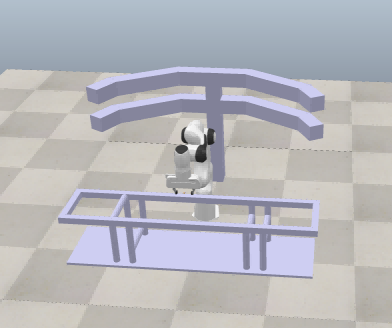}}
    \caption{10 typical scenes used for evaluating ViIK in this paper which are proposed in \cite{benchmark}. (Env-1, Env-2), (Env-3, Env-4), (Env-5, Env-6), (Env-7, Env-8), and (Env-9, Env-10) are used for the model, ViIK-2. Env-1 -- Env-5, and Env-6 -- Env-10 are used for the model, ViIK-5. Env-1 -- Env-10 are used for the model, ViIK-10. }
    \label{fig.benchmark}
\end{figure*}


\section*{APPENDIX}

\subsection{10 scenes used in this paper}\label{sec.app_proof}
We present the scenes used in this paper in Fig. \ref{fig.benchmark}. All scenes are chosen from the benchmark proposed in\cite{benchmark}.

\subsection{Experimental details}\label{sec.app_exp}
\begin{figure}[h]
	\centering
	\subfloat{
		\includegraphics[width=0.1\textwidth]{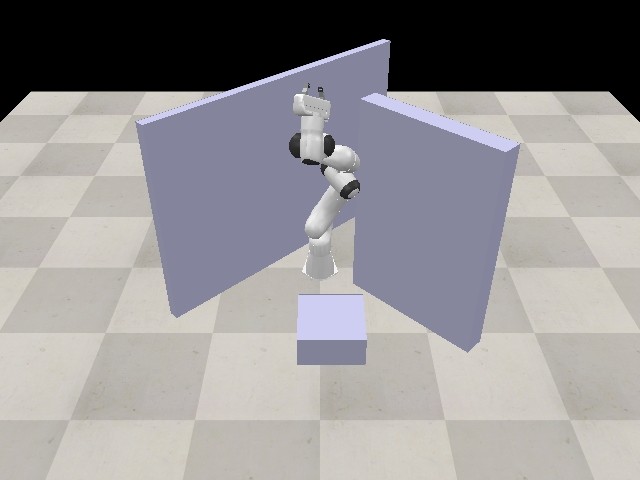}
        \includegraphics[width=0.1\textwidth]{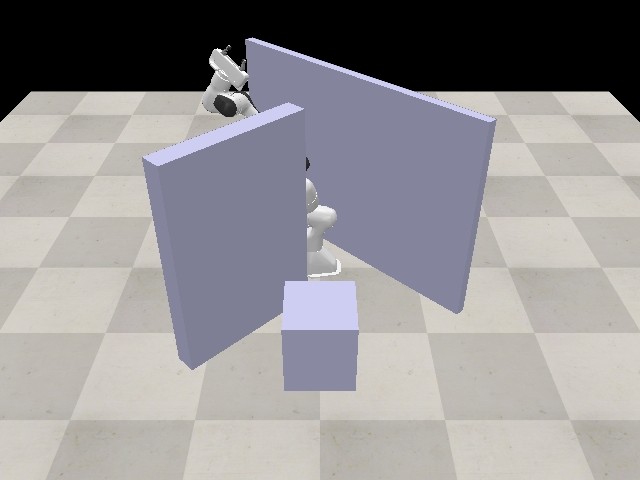}
        \includegraphics[width=0.1\textwidth]{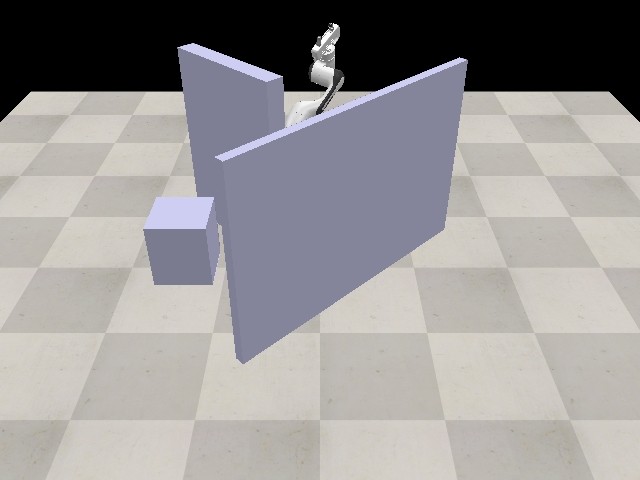}
        \includegraphics[width=0.1\textwidth]{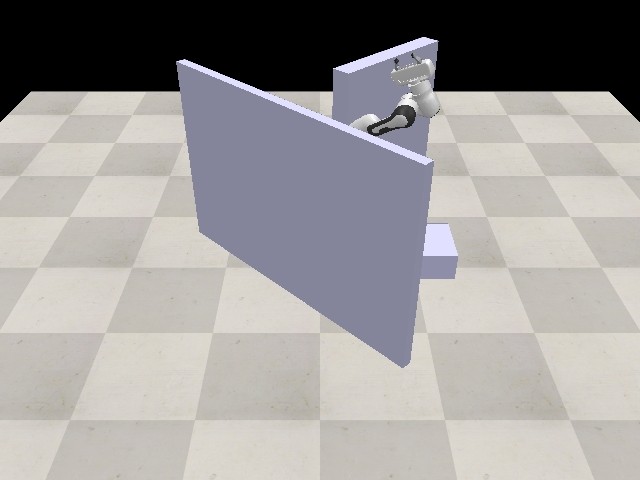}
        }\\[3pt]
    \caption{An example of multi-view images used in this paper. }
    \label{fig.multi-views}
\end{figure}

All models are GLOW-based\cite{glow} with 24 coupling layers. The training set has 5 million random samples in the C-space with the corresponding end-effector poses and 4882 sets of multi-view images of each environment. The batch size is 1024. The Adamw optimizer is used for training all models. The hyperparameters for training ViIK are shown in Tab. \ref{tab.hyper}. Moreover, the multi-views used in this paper are presented in Fig. \ref{fig.multi-views} with an example in Env-1. 

\begin{table}[h]
    \caption{Hyperparameters for training ViIK.}
    \centering
    \begin{threeparttable}
    \begin{tabular}{lcccc}
        \toprule [1pt]\noalign{\vskip 2pt}
        model & NF Blocks & Learning Rate & Gamma & Epochs\\[5pt]
        \hline\noalign{\vskip 5pt}
        ViIK-2   & 24    & $3\times10^{-5}$   & 0.988553095 & 700  \\
        ViIK-5   & 36    & $2\times10^{-5}$   & 0.992354096 & 1000  \\
        ViIK-10  & 48    & $1.5\times10^{-5}$ & 0.994260074 & 1300 \\[5pt]
        \bottomrule [1pt]
    \end{tabular}
    \vspace{5pt}
    \end{threeparttable}
\label{tab.hyper}
\end{table}

\begin{thebibliography}{1}
\bibitem{quickhull}
C. B. Barber, D. P. Dobkin, and H. Huhdanpaa, "The quickhull algorithm for convex hulls," ACM Trans. Math. Softw. 22, 4 (Dec. 1996), 469–483. https://doi.org/10.1145/235815.235821
\bibitem{ikfast}
R. Diankov, “Automated construction of robotic manipulation programs,” Ph.D. dissertation, School Comput. Sci., Carnegie Mellon Univ., Pittsburgh, PA, USA, 2010.
\bibitem{ikbt}
D. Zhang and B. Hannaford, “IKBT: Solving closed-form inverse kinematics with behavior tree,” J. Artif. Intell. Res., vol. 65, pp. 457–486, Nov. 2017.
\bibitem{stampede}
D. Rakita, B. Mutlu, and M. Gleicher, “STAMPEDE: A discrete optimization method for solving pathwise-inverse kinematics,” in Proc. Int. Conf. Robot. Automat., 2019, pp. 3507–3513.
\bibitem{trac-ik}
P. Beeson and B. Ames, “TRAC-IK: An open-source library for improved solving of generic inverse kinematics,” in Proc. IEEE-RAS 15th Int. Conf. Humanoid Robots, 2015, pp. 928–935.
\bibitem{nn-ik-1}
R. J. Ahmed, L. C. A. Dülger, and S. Kapucu, “A new artificial neural
network approach in solving inverse kinematics of robotic arm (densoVP6242),” Comput. Intell. Neurosci., vol. 2016, 2016, Art. no. 5720163. [Online]. Available: https://dl.acm.org/doi/10.1155/2016/5720163
\bibitem{nn-ik-2}
A. Csiszar, J. Eilers, and A. Verl, “On solving the inverse kinematics problem using neural networks,” in Proc. 24th Int. Conf. Mechatronics Mach. Vis. Pract., 2017, pp. 1–6.
\bibitem{nn-ik-3}
J. Demby’s, Y. Gao, and G. N. Desouza, “A study on solving the inverse kinematics of serial robots using artificial neural network and fuzzy neural network,” in Proc. IEEE Int. Conf. Fuzzy Syst., 2019, pp. 1–6.
\bibitem{nn-ik-4}
H. Ren and P. Ben-Tzvi, “Learning inverse kinematics and dynamics of a robotic manipulator using generative adversarial networks,” Robot. Auton. Syst., vol. 124, Feb. 2020, Art. no. 103386.
\bibitem{nn-ik-5}
S. Kim and J. Perez, “Learning reachable manifold and inverse mapping for a redundant robot manipulator,” in Proc. Int. Conf. Robot. Automat., 2021, pp. 4731–4737.
\bibitem{ikflow-1}
L. Ardizzone et al., “Analyzing inverse problems with invertible neural networks,” in Proc. Int. Conf. Representation Learn., 2019. [Online]. Available: https://arxiv.org/abs/1808.04730v3
\bibitem{ikflow-2}
B. Ames, J. Morgan and G. Konidaris, "IKFlow: Generating Diverse Inverse Kinematics Solutions," in IEEE Robotics and Automation Letters, vol. 7, no. 3, pp. 7177-7184, July 2022, doi: 10.1109/LRA.2022.3181374.
\bibitem{paddingflow}
Q. Meng, C. Xia and X. Wang, "PaddingFlow: Improving Normalizing Flows with Padding-Dimensional Noise," in arXiv preprint arXiv:2403.08216, 2024.
\bibitem{discrete}
B. Uria, I. Murray and H. Larochelle, “RNADE: The real-valued neural autoregressive density-estimator,” in Advances in Neural Information Processing Systems, 2013.
\bibitem{softflow}
H. Kim, H. Lee, W. H. Kang, J. Y. Lee, and N. S. Kim, “SoftFlow: Probabilistic Framework for Normalizing Flow on Manifolds,” in Advances in Neural Information Processing Systems, 2020.
\bibitem{survey_nf}
G. Papamakarios, E. Nalisnick, D. J. Rezende, S. Mohamed and B. Lakshminarayanan, “Normalizing Flows for Probabilistic Modeling and Inference,” in Journal of Machine Learning Research, vol. 22, no. 57, pp. 1-64, 2021.
\bibitem{benchmark}
C. Gaebert, S. Kaden, B. Fischer and U. Thomas, "Parameter Optimization for Manipulator Motion Planning using a Novel Benchmark Set," 2023 IEEE International Conference on Robotics and Automation (ICRA), London, United Kingdom, 2023, pp. 9218-9223, doi: 10.1109/ICRA48891.2023.10160694.
\bibitem{glow}
P. D. Kingma and P. Dhariwal, “Glow: Generative flow with invertible 1 × 1 convolutions,” in Proc. 32nd Conf. Neural Inf. Process. Syst., 2018. [Online]. Available: https://github.com/openai/glow
\end{thebibliography}
\end{document}